# Classifying the Unstructured IT Service Desk Tickets Using Ensemble of Classifiers

Paramesh S.P
Department of CS & E
U.B.D.T College of Engineering
Davanagere, Karnataka, India
sp.paramesh@gmail.com

Ramya C
Department of CS & E
U.B.D.T College of Engineering
Davanagere, Karnataka, India
cramyac@gmail.com

Dr. Shreedhara K.S
Department of CS & E
U.B.D.T College of Engineering
Davanagere , Karnataka,India
ks_shreedhara@yahoo.com

*Abstract*— Manual classification of IT service desk tickets may result in routing of the tickets to the wrong resolution group. Incorrect assignment of IT service desk tickets leads to reassignment of tickets, unnecessary resource utilization and delays the resolution time. Traditional machine learning algorithms can be used to automatically classify the IT service desk tickets. Service desk ticket classifier models can be trained by mining the historical unstructured ticket description and the corresponding label. The model can then be used to classify the new service desk ticket based on the ticket description. The performance of the traditional classifier systems can be further improved by using various ensemble of classification techniques. This paper brings out the three most popular ensemble methods ie, Bagging, Boosting and Voting ensemble for combining the predictions from different models to further improve the accuracy of the ticket classifier system. The performance of the ensemble classifier system is checked against the individual base classifiers using various performance metrics. Ensemble of classifiers performed well in comparison with the corresponding base classifiers. The advantages of building such an automated ticket classifier systems are simplified user interface, faster resolution time, improved productivity, customer satisfaction and growth in business. The real world service desk ticket data from a large enterprise IT infrastructure is used for our research purpose.

## I. INTRODUCTION

In a typical IT organization, employee's experiences lot of issues with respect to facilities, Infrastructure, applications, HR related issues, travel etc. The employees of the organization raise the issue ticket through the IT service desk or Helpdesk which is typically of web based. The submitted tickets will then be assigned to a proper domain expert group or service desk agent for resolving the problem ticket depending upon the category of the ticket. Web based IT service desk systems contains the structured fields like submitter, ticket category, priority, severity etc. It also contains the free form fields like ticket description where the user enters the description about the ticket in his natural language. The structure of the typical IT service desk tickets raised by employees of the organization is given in Table 1.

TABLE 1. TYPICAL SERVICE DESK TICKET DATA

| Ticket ID | Ticket category | priority | submitter | Ticket description | Status |
|---|---|---|---|---|---|
| 100 | Network issue | High | XYZ | Unable to login into LAN | Open |
| 101 | Hardware Problem | Medium | ABC | Hard disk crashed | In progress |
| 103 | Software Install | High | PQR | Need to install Eclipse | Open |

While creating the problem tickets, employees manually choose the category of the problem, priority, severity of the issues and also enter the description of the ticket in natural language. Manual selection of the ticket category by end user may results in wrong categorization of the ticket since it is based on the users perception of the problem and whether the user chose the correct category to log the problem. Wrongly categorized tickets will be assigned to the incorrect resolution group and results in delay in obtaining the solution for the given problem tickets. The existing service desk systems work well for structured data. To overcome all these problems, we can build an automated ticket classifier system using various machine learning techniques. Such an automated ticket classifier categorises the service desk ticket by analysing the natural language ticket description entered by end user. Ticket classifier models can be developed by using both supervised and unsupervised machine learning techniques. When the label or category of the historical ticket data is known in prior then supervised machine learning techniques like classification algorithms can be used to build classifier models [1], [5], [7], [8] and [9]. Unsupervised machine learning techniques like clustering can also be used to group similar kind of tickets followed by labelling of the clusters when the ticket category is unknown [4], [6] and [11].

The development of such an automated ticket classifier system for one of the real world enterprise IT infrastructure service desk ticket data is discussed in Paramesh et.al [1] by same author.The work in [1] uses various supervised machine learning techniques like Logistic regression, Multinomial Naïve Bayes, K-nearest neighbour and Support vector machines to build classifier models. The user needs to specify only the natural language ticket description which is unstructured in nature. The ticket classifier model automatically classifies the ticket into one of the predefined category using the user's ticket description and routes the ticket to proper domain team for resolution. Historical dataset containing description of ticket and corresponding label or category is used to train the classifier. Dataset may also contain other structured fields but only the free form description field and the fixed category fields are taken into consideration for building the ticket classifier model.

As a future work of [1], in this research paper we use an ensemble of classifier models to further improve the accuracy of the classifier system [22]. Ensemble involves combining the results of multiple models. The result of combining different models gives better accuracy when compared to the individual models [18].Some of the ensemble methods such as bagging, boosting and simple voting ensemble classifier are used for the ticket classification purposes. In this research work, Bagging techniques [19] are applied for individual base classifiers such as Decision tree, Multinomial Naïve Bayes (MNB) and Support Vector Machines (SVM). RandomForest classifier models are also built as a part of bagging methods. Boosting techniques such as Adaboost [20] are used in this research work. A combination of logistic regression (LR), SVM, MNB and K-Nearest Neighbour (KNN) are used as an input to the Voting ensemble classifier. The performance of all the ensemble models are evaluated against individual classifier models using various classification performance metrics. In this work, service desk data of one of the real world enterprise IT infrastructure is used for research purposes. Typical IT infrastructure problems can be related to hardware issues, software issues, network issues, email issues etc.

Ticket classification is a use case of text document classification in which each ticket description is considered as one single document and the category of the ticket is the label of the document. So to build ticket classifier model, it follows all the phases of a typical text classification problem ie, data pre-processing, feature selection, feature vector representation followed by model building and evaluation [14]-[16].The IT infrastructure ticket data considered in this research work had huge amount of unwanted and unclean data and hence handling data related challenges is one of the key issues in developing such a ticket classifier. Main objectives of this paper are-

- Building the IT service desk ticket classifiers using various Ensemble of classifier methods such as Bagging, Boosting and voting ensemble methods.
- Evaluation and comparison of various ensemble classifiers with the traditional base classifiers using various classification performance evaluation metrics.

The advantages of developing such an automated ticket classifier systems are simplified user interface, faster resolution time,effective resource utilization, improvement in customer satisfaction and growth in business.

## II. LITERATURE SURVEY

Some of the prior researches in the field of automation of service desk systems are given below.

Paramesh S.P et al. [1] proposed a method for building an automated service desk ticket classifier system by considering the IT infrastructure helpdesk data.Traditional supervised machine learning techniques like Logistic regression, KNN, MNB and SVM are used to build classifier models. The author also discusses various techniques used to handle the data related challenges during the pre-processing phase. Techniques for handling unwanted data, imbalanced data and wrongly labelled data are discussed in detail. Out of all the different classification models considered, SVM outperformed well when compared to other models.

S. Agarwal et al [2] discusses about building a system that extracts knowledge about different categories of IT infrastructure problems and to provide the automatic resolution for the new issue tickets. The system uses a historical problem tickets and the corresponding resolution data to achieve the task. Machine learning and natural language processing techniques are used to develop such system.

C. Zeng et al [3] uses hierarchical multi-label classification to classify the monitoring tickets. The proposed method uses the novel GLabel algorithm to identify the optimal hierarchical multi-labels for each ticket instance. During the mining process, the knowledge of the domain experts is used to further improve the classification results.

S. Roy et al [4] propose a model to cluster and label the incident tickets using the user's ticket description. The method uses the unsupervised machine learning techniques like clustering to categorize the user tickets. The proposed method applies new distance metric which uses the combination of Jaccard distance and cosine distance for fixed and free fields of the tickets respectively. The distance metric is used to computer the distance between the tickets. K-means clustering uses this distance metric to find the different ticket clusters which is followed by labelling of the clusters. To find the cluster labels, the method first extracts the logical item sets for each cluster using the modified Logical Item set mining and then does the semantic labelling for each item sets of the cluster.

Jian Xu et al [5] use STI-E model, an ensemble of SVMs to classify situation tickets from the historical data. In this model, a domain word discovery algorithm is used to extract domain word list and a selective labelling policy is then used to select training data using the extracted key words. Finally, SVM ensemble classifier is used to identify the situation tickets. The SVM ensemble performed well when compared to the base SVM classifier.

In [6] probabilistic concept model is proposed to examine the service desk tickets. The framework uses combination of topic modelling, clustering and IR techniques to analyse the tickets. This approach extracts concepts from the phrases of the ticket description and then clusters the tickets using these concepts.

Gargi et al [7], proposed a ticket classifier model called BlueFin for categorization of messy and unstructured problem tickets. The approach involves correlation of event and ticket data followed by the context based classification of correlated data. Performance of Bluefin is compared against another ticket classifier tool called SmartDispatch based on SVM. Bluefin performed extremely well for most of the samples.

C. Kadar et al. [8] proposed an approach that classifies the change requests (CR) into one of the activities in a catalogue. It uses Information Retrieval (IR) approach using Lucene, an open source text search engine and supervised machine learning techniques for change requests classification. Machine learning techniques performed well when compared to IR approach but requires large training data. So to speed up the learning, techniques like active learning approaches were used.

S. Agarwal et al. [9] proposed an automated ticket classifier system called Smart dispatch based on a combination of SVM classification and a discriminative term weighting schemes. The tool does the ticket classification by mining the old user ticket description and the associated label.

R.Potharaju et al. [10] proposes a system called NetSieve which analyses the network trouble tickets using the ticket description entered by user to infer the problem symptoms, trouble shooting activities and resolution actions. In order to achieve these goals, NetSieve combines the NLP techniques, knowledge representation and Ontology modelling.

Shimpi et al. [11] proposes a model to extract issues from the ticket descriptions. Information retrieval techniques along with domain knowledge are used to extract the issues form the problem tickets. The author uses clustering based techniques and keyword search based approaches for extracting issues from system and user generated tickets respectively.

Our research problem is a use case of text document classification problem. Sebastiani has given an excellent review on text classification problem [13].The paper discusses various dimensionality reduction techniques for feature selection and text classifiers for text categorization.

Li and Jain [14], discusses the document classification problem using some individual base classifiers and combination of base classifiers. Comparative study of the performance of the different classifier models is also discussed.

M. Ikonomakis et al. [15] discusses the various phases involved in the text classification process. The techniques required to achieve data pre-processing, document representation, feature selection, feature transformation and to build and evaluate the classification models are discussed in detail.

A.Khan et al [16] gives an excellent review on the machine learning techniques used for the classification of text documents. Review on the approaches used for the Document representation, feature extraction and feature selection are also discussed.

The dataset considered for our research work had some imbalanced data with some categories containing more number of instances and some with less number of instances. Class imbalance problems in a multiclass problem affect the performances of the classifier models. Sotiris in [17] discusses various methods to handle such class imbalance problems. Sampling methods like random under sampling and oversampling were used to balance the imbalance dataset.

Ensemble classifiers combine the predictions from multiple classifiers to improve the accuracy and the execution time of the classifier models. In the literature, different types of ensemble techniques such as bagging [19] and boosting [20] were discussed in the context of classification problems.

Luis et al [21] proposes an ensemble methodology based on the mean co-association matrix for the text classification task. Experiments reveal that the proposed method performs well when compared to the individual base classifier considered for the research.

R.N. Behera et al [22] gives a review on the ensemble machine learning approaches such as bagging and boosting used in the classification of human sentiments.

## III. PROPOSED METHODOLOGY

Service desk ticket classification can be considered as a text document classification.To develop a ticket classifier system, the historical ticket data must undergo data pre-processing to remove any unwanted and noisy data followed by representation of the ticket data. The classification algorithms are then applied to training data to build classifier models. The high level components of the proposed service desk ticket classifier system are given in Fig 1.

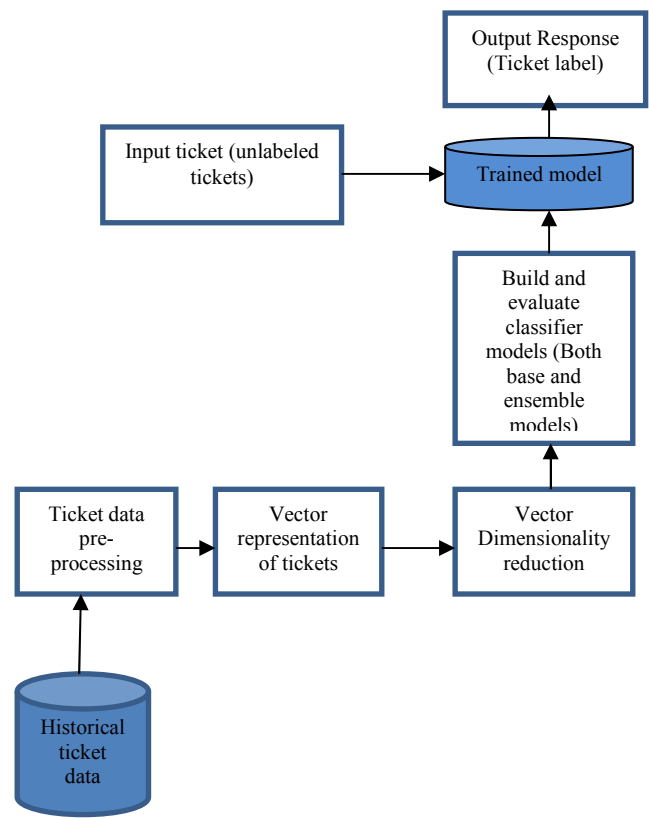

Fig.1. Solution diagram for IT service desk ticket classifier

The main components of the proposed system are explained as follows.

### A. Data pre-processing

The pre-processing of training data is one of the important steps in the data mining process. Pre-processing usually involves cleaning of the raw data ie, in our case it is the cleaning of the historical ticket data. The dataset should be clean without any noisy and unwanted data in order to build accurate machine learning models. The ticket data used in this work had lot of unwanted data such as names, email address,phone numbers, special characters, date, time, etc. The pre-processing block cleans all such undesired data since they do not aid in ticket classification. Commonly used stop words and functional words are also removed as a part of pre-processing. The dataset also had imbalanced data, so techniques like random oversampling and random under sampling are used to balance such data from the dataset[17]. The challenges faced and the technical steps in pre-processing of such a noisy unstructured ticket data are detailed in [1] by same author.

### B. Feature Vector representation of ticket data and Dimensionality reduction

Historical ticket descriptions (training data) must be represented in numerical form before applying any machine learning algorithms. A Feature vector is constructed on each ticket description using Term Frequency-Inverse document frequency(TF-IDF) weighting scheme. In this representation, each element in the vector specifies the unique words taken from the entire corpus and is assigned a TF-IDF value. Once the ticket data is represented using feature vector representation, feature selection techniques such as chi square test is applied to reduce the number of features[16].

### C. Building Classification models

In this work, popular ensemble of classifier methods such as Bagging, Boosting and Voting methods are used to build ticket classifiers.The performance of both base classifier models and ensemble classifier models is evaluated and compared using various evaluation metrics such as Accuracy, Precision, Recall and F1-score. At the end of this phase, the model which performs best is chosen and is stored in the model store for the classification of new ticket instance.

*1) Bagging (Bootstrap Aggregation)* :

Bagging involves creating n number of models of the same classifier using n different subsamples (chosen with replacement) of the training data.The classifier models are developed in parallel and the final output prediction is the aggregation or averaging of each individual predictions. Bagging is also called Bootstrap Aggregation [19]. In this paper, we used the following bagging models for building the service desk ticket classifier system.

   a) *Bagged Decision trees*
   b) *Bagged MNB*
   c) *Bagged SVM*
   d) *Random Forest classifier*

Performance of all the Bagged classifier models is compared with their corresponding base classifiers models using various performance evaluation metrics.

*2) Boosting:*

Boosting involves creating a sequence of models which tries to correct the mistakes of the previous models in the sequence[22]. If an instance was misclassified, it tries to increase the weight of this instance and vice versa. Generally, the boosting techniques are used to boost the performance of weak learners[20]. In this research work, we used the Adaboost classifier algorithm to build the boosting model for the weak classifier such as Decision Stump, a type of decision tree with depth=1.

*3) Voting Ensemble:*

Voting Ensemble involves combining the predictions from multiple machine learning algorithms. It builds the sub models using various base classifiers and then averages the predictions of all base classifiers to get the prediction of the new unseen data. In this research, voting ensemble classifier is applied on the individual base models like logistic regression (LR) SVM, MNB and KNN.

### D. Trained model

Accuracy performances of different ensemble models are evaluated and compared with base classifier model. After the comparative study, the best trained model is selected and is stored inside this model store. When a new unlabelled ticket arrives, the ticket category is automatically predicted using the classifier model stored in the model store.

## IV. IMPLEMENTATION AND EXPERIMENTAL RESULTS

For the implementation purpose of this research topic, we used the Python as programming language with all necessary libraries such as pandas, scikit learn, matplotlib etc.

### A. Data set analysis and pre-processing

We used service desk ticket data from a large enterprise IT infrastructure (collected over a period of one month) for our research purpose. Each ticket contains many fixed fields like ticket category, priority etc and free form fields like ticket description. Only ticket description and its category were used to build the classifier models and all other fields were ignored.

Initial analysis of the data revealed the following details.

- Total number of tickets collected: 10742 instances.
- No of distinct classes present in the dataset: 18 classes (hence it multi-class problem).

The view of the tickets distribution across multiple classes is given in Fig 2.

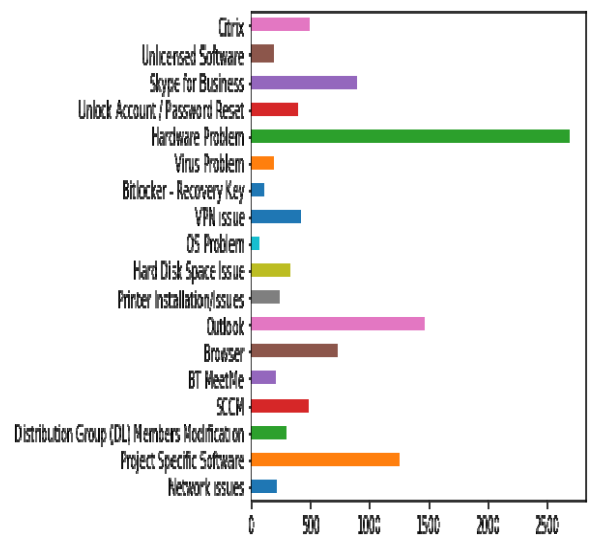

Fig.2. Ticket Distribution across different classes

Careful data analysis also reveals that, huge amount of unwanted text, incorrect labelled data (noisy label) and imbalanced classes were present in the raw data. Details of the methods used for pre-processing of unstructured ticket data are discussed in detail in Paramesh et.al [1] by same author. After the pre-processing of the ticket data, numerical representation of the ticket data is done using the vector space model followed by dimensionality reduction using chi square method.

*B. Building and Evaluation of the Ensemble Classifier models*

In order to build different ensemble of classifiers from the base classifier models, the pre-processed data would be split into training and test sets [13]. We used 70:30 percentage split ratio with 70% of tickets used as training set and rest for testing the classification accuracy. The classifier models are built by using various ensemble models such as bagging, boosting and voting ensemble methods. For performance evaluation of classifiers, we used metrics like accuracy, precision, recall and F-score.

*1) Bagging models :*
We experimented the below bagging models on our training ticket data.

- Bagging- Decision tree (Bagging-Dtree)
- Bagging –Multinomial Naïve Bayes(Bagging-MNB)
- Bagging Support Vector Machines(Bagging-SVM)
- Random Forest classifier.

The comparative study of accuracy performance of bagging classifiers with that of the corresponding base classifiers is given in Table 2 and Fig 3.

TABLE 2. PERFORMANCE OF BAGGED CLASSIFIERS AND THE CORRESPONDING BASE CLASSIFIER

| Classifier model | Accuracy | Precision | Recall | F-score |
|---|---|---|---|---|
| DTree | 0.906003 | 0.907807 | 0.906003 | 0.905942 |
| Bagging-Dtree | 0.920428 | 0.922252 | 0.920428 | 0.920530 |
| MNB | 0.699395 | 0.760604 | 0.699395 | 0.672432 |
| Bagging-MNB | 0.701722 | 0.763395 | 0.701722 | 0.674518 |
| RandomForest | 0.918101 | 0.919531 | 0.918101 | 0.918060 |
| SVM | 0.882271 | 0.882717 | 0.882271 | 0.881410 |
| Bagging-SVM | 0.882271 | 0.882717 | 0.882271 | 0.881410 |

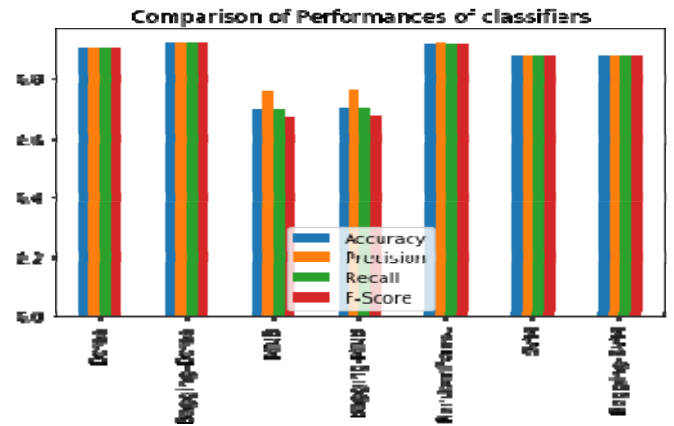

Fig.3. Performance comparison of Bagged Classifiers and the corresponding base classifier

Above results indicates that the performance of bagged classifier model is higher when compared to individual base models. For example, the accuracy of Decision tree model and the Bagged- decision tree model is 90.60% and 92.04% respectively. Similarly the accuracy of Multinomial Naïve Bayes (MNB) and the Bagged- MNB model is 69.93% and 70.17% respectively. The accuracy of RandomForest classifier (an extension of Bagged decision tree classifier) is 91.81% which is higher than the accuracy of base Decision tree classifier (90.60%).

*2) Boosting*
We used the Adaboost classifier to build the boosting model. Adaboost classifier model is used to boost the performance of weak learners such as Decision Stump (a type of decision tree with depth=1). The comparative study of accuracy performance of Adaboost classifier and the Decision Stump classifier is given in Table 3 and Fig 4.

TABLE 3: PERFORMANCES OF ADABOOST CLASSIFIER AND THE DECISION STUMP

| Classifier | Accuracy | Precision | Recall | f-score |
|---|---|---|---|---|
| Decision stump tree | 0.344812 | 0.195864 | 0.344812 | 0.217571 |
| Adaboost classifier | 0.535598 | 0.458315 | 0.535598 | 0.461304 |

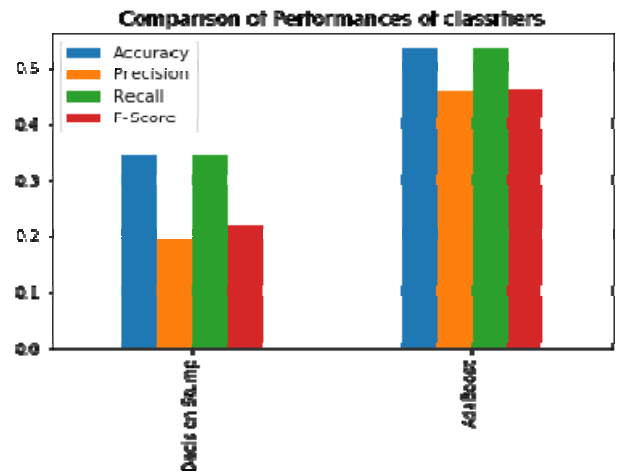

Fig.4. Performance comparison of Adaboost Classifier and the Decision stump tree

From the above results, we could see that Adaboost model boosted the performance of Decision stump classifier.

*3) Voting ensemble classifier*

In our research, Voting classifier is applied on the below base classifier models.

- Logistic regression (LR)
- K-Nearest-Neighbour(KNN) (with K=5)
- Multinomial Naïve Bayes (MNB)
- Support vector machines (SVM)

The accuracy performance of the Voting classifier and the base classifier models on our dataset is given in Table 4 and Fig 5.

The accuracy performance of LR, KNN, MNB and SVM are respectively 80.50%, 69.24%, 70.1% and 88.22% respectively. The voting classifier gives accuracy of 81.11% which is the mean or average of all the base classifiers.

TABLE 4: PERFORMANCES OF VOTING CLASSIFIER AND THE BASE CLASSIFIERS

| Classifier | Accuracy | Precision | Recall | F-score |
|---|---|---|---|---|
| Logistic Regression (LR) | 0.805026 | 0.820461 | 0.805026 | 0.802369 |
| KNN | 0.692415 | 0.747543 | 0.692415 | 0.696177 |
| MNB | 0.701722 | 0.763395 | 0.701722 | 0.674518 |
| SVM | 0.882271 | 0.882717 | 0.882271 | 0.881410 |
| Voting-Classifier | 0.811540 | 0.829686 | 0.811540 | 0.809981 |

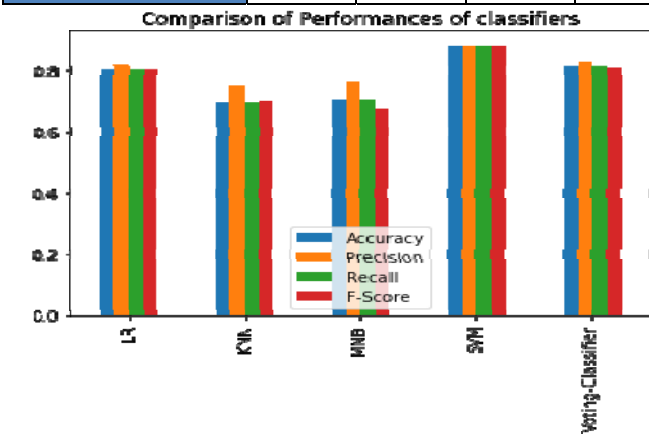

Fig.5. Performances comparison of Voting classifier and the base classifiers

## V. CONCLUSION

Assignment of service desk tickets to a proper domain expert team or resolution team is a crucial step in any helpdesk systems. Incorrect routing of tickets to wrong resolver group results in delay in obtaining the solution to the problem, unnecessary resource utilization, customer satisfaction deterioration and affects the business. To overcome all these problems, we proposed an automated ticket classifier system based on supervised machine learning techniques which automatically predicts the category of the ticket using the natural language ticket description entered by user. In this research work,we developed ticket classifier model for one of the real time IT infrastructure service desk. We used ensemble of classification methods such as Bagging, Boosting and voting ensemble methods along with some base classifier models to build the classifiers. The performance of all the ensemble classifiers is evaluated and compared with the corresponding base classifiers using various classifier performance metrics.Ensemble classifiers outperformed well in comparison to the base classifiers. Since the user's natural language ticket description is highly unstructured in nature, different data cleaning techniques are used handle the noise present in the ticket data. Techniques like random under sampling and random oversampling is used to handle class imbalance problems found in the initial training data. The proposed ticket classifier system results in effective support resource utilization, improved end user experience, and quicker turnaround time.

## VI. FUTURE ENHANCEMENTS

The domain specific keywords and stop words could be extracted from the training data to ensure only relevant features will be used for building the models to further improve the accuracy of the system.We plan to explore Deep learning based classification models to automatically classify the service desk tickets and to investigate its performance on our IT infrastructure ticket data. We plan to consider service desk data from different domains with even larger training dataset and multiple classes.

ACKNOWLEDGEMENT

I give my heartfelt thanks to Computer science research centre UBDTCE, Davanagere that has enabled me to build this system. I would like to thank my Ph.D research guide Dr.Shreedhara K..S under whose able guidance this research work has been carried out and completed successfully. I would like to thank all my family members for their kind support.